\title{emnlp 2017 temporal camera ready}
\documentclass[11pt,letterpaper]{article}
\usepackage{emnlp2017}
\usepackage{times}
\usepackage{latexsym}
\usepackage{graphicx}

\emnlpfinalcopy

\def\emnlppaperid{***}

\title{A Sequential Model for Classifying Temporal Relations between Intra-Sentence Events}

\author {Prafulla Kumar Choubey \and Ruihong Huang \\
         Department of Computer Science and Engineering\\
		Texas A\&M University\\
         {\tt (prafulla.choubey, huangrh)@tamu.edu}}

\date{}

\begin{document}

\maketitle

\begin{abstract}
  We present a sequential model for temporal relation classification between intra-sentence events. The key observation is that the overall syntactic structure and compositional meanings of the multi-word context between events are important for distinguishing among fine-grained temporal relations. Specifically, our approach first extracts a sequence of context words that indicates the temporal relation between two events, which well align with the dependency path between two event mentions. The context word sequence, together with a parts-of-speech tag sequence and a dependency relation sequence that are generated corresponding to the word sequence, are then provided as input to bidirectional recurrent neural network (LSTM) models. The neural nets learn compositional syntactic and semantic representations of contexts surrounding the two events and predict the temporal relation between them. Evaluation of the proposed approach on TimeBank corpus shows that sequential modeling is capable of accurately recognizing temporal relations between events, which outperforms a neural net model using various discrete features as input that imitates previous feature based models.
\end{abstract}

\section{Introduction}

Identifying temporal relations between events is crucial to constructing events timeline. It has direct application in tasks such as question answering, event timeline generation and document summarization. 

Previous works studied this task as the classification problem based on discrete features defined over lexico-syntactic, semantic and discourse features. However, these features are often derived from local contexts of  two events and are only capable of capturing direct evidences indicating the temporal relation. Specifically, when two events are distantly located or are separated by other events in between, feature based approaches often fail to utilize  compositional evidences, which are hard to encode using discrete features.

\begin{figure}
\noindent\fbox{%
\parbox{0.465\textwidth}{%
\vspace*{1ex}
Bush said he saw little reason to be optimistic about a settlement of the {\bf dispute}, which stems from Iraq's {\bf invasion} of oil-wealthy Kuwait and its subsequent military {\bf buildup} on the border of Saudi Arabia.  

Relations: {\bf (}dispute $after\ ^{rel_1}$ invasion, invasion $ibefore\ ^{rel_2}$ buildup, dispute $after\ ^{rel_3}$ buildup{\bf )}

\centering \includegraphics [width=2.9in]{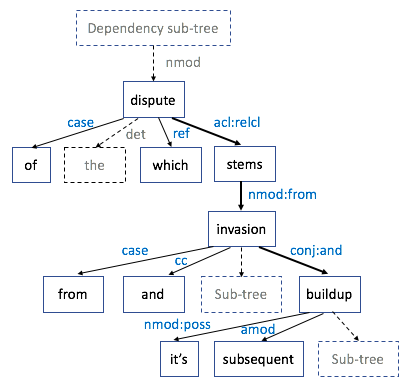}
}
} 
\caption{\label{example} Example sentence to illustrate the temporal context for event pairs. }
\end{figure}

Consider the example sentence in Figure~\ref{example}. Here, the first two temporal relations, $dispute\  {\bf after\ ^{rel_1}} \ invasion$ and $invation\ {\bf ibefore\ ^{rel_2}}\ buildup$, involve events that are close by and discrete features, such as dependency relations and bag-of-words extracted from local contexts of two events, might be sufficient to correctly detect their relations. However, for the temporal relation $dispute\ {\bf after\ ^{rel_3}}\ buildup$, 
the context between the two events is long, complex and involves another event ($\it invasion$) as well, which makes it challenging for any individual feature or feature combinations to capture the temporal relation.

We propose that the overall syntactic structure of in-between contexts including the linear order of words as well as the compositional  semantics of multi-word contexts are critical for predicting the temporal relation between two events. Furthermore, the most important syntactic and semantic structures are derived along dependency paths between two event mentions\footnote{In this paper, we restrict ourselves to study temporal relation classification between event mentions that are within one sentence.}. This aligns well with the observation that semantic composition relates to grammatical dependency relations~\cite{monroe2014dependency,reddy2016transforming}. 

Our approach defines rules on dependency parse trees to extract temporal relation indicating contexts. First, we extract the dependency path between two event mentions. Then we apply two heuristic rules to enrich extracted dependency paths and deal with complex syntactic structures such as punctuations. Empirically, we found that parts-of-speech tags (POS) and dependency sequences generated following the dependency path provide evidences to predict the temporal relation as well. 

We use neural net sequence models to capture structural and semantic compositionality in describing temporal relations between events.  Specifically, we generate three sequences for each dependency path, the word sequence, the POS tag sequence and the dependency relation sequence. Using the three types of sequences as input, we train bi-directional LSTM models that consume each of the three sequences and model compositional structural information, both syntactically and semantically. 

The evaluation shows that each type of sequences is useful to temporal relation classification between events. Our complete neural net model taking all the three types of sequences performs the best, which clearly outperforms feature based models.

\section{Related Works}
Most of the previous works on temporal relation classification are based on feature-based classifiers.~\citet{mani2006machine} built MaxEnt classifier on hand-tagged features in the corpus, including tense, aspect, modality, polarity and event class for classifying temporal relations. Later ~\citet{chambers2007classifying} used a two-stage classifier which first learned imperfect event attributes and then combined them with other linguistic features in the second stage to perform the classification.

The following works mostly expanded the feature sets~\cite{cheng2007naist, bethard2007cu,uzzaman2012tempeval,bethard2013cleartk,kolomiyets2012extracting,chambers2013navytime,laokulrat2013uttime}. Specifically, ~\citet{chambers2013navytime} used direct dependency path between event pairs to capture syntactic context.  ~\citet{laokulrat2013uttime} used 3-grams of paths between two event mentions in a dependency tree as features instead of full paths as those are too sparse. We found that modeling the entire path as one sequence provides greater compositional evidence on the temporal relation. In addition, modifiers attached to the words in a path with specific dependency relations like {\it nmod:tmod} are also informative.

\citet{d2013classifying} proposed a hybrid system for temporal relation classification that combines the learned classifier with 437 hand-coded rules. Their system first applied high-accuracy rules and then used the learned classifier, trained on rich features including those high-accuracy rules as features, to classify the cases that were not handled by the rules. ~\citet{ng2013exploiting} also showed the effectiveness of different discourse analysis frameworks for this task.
Later~\citet{mirza2014classifying} showed that a simpler approach based on lexico-syntactic features achieved results comparable to~\citet{d2013classifying}. They also reported that dependency order between events, either governor-dependent
or dependent-governor, was not useful in their experiments. However, we show that dependency relations, when modeled as a sequence, contribute significantly to this task. 

\section{Temporal Link Labeling}
In this section, we describe the task of temporal relation classification, dataset, context words sequence extraction model and the used recurrent neural net based classifier.

\subsection{Task description}
Early works on temporal relation classification ~\citet{mani2006machine,chambers2007classifying} and the first two versions of TempEval ~\cite{verhagen2007semeval,verhagen2010semeval} simplified the task by 
considering only six relation types. They combined the pair of relation types that are the inverse of each other and ignored the relations {\it during} and {\it during\_inv}. Then TempEval-3~\cite{Uzzaman_semeval-2013task} extended the task to complete 14 class classification problem and all later works have considered all 14 relations. Our model performs 14-class classification following the recent works, as this is arguably more challenging ~\cite{d2013classifying}.
Also, we consider gold annotated event pairs, mainly because the corpus is small and distribution of relations is very skewed. All previous works focusing on the problem of classifying temporal relation types assumed gold annotation.

\subsection{Dataset}

\begin{table}[h]
\small
\begin{center}
\begin{tabular}{|l|l|l|l|}
\hline \bf Relations & \bf Train  & \bf Validate  & \bf Test \\ \hline
After & 419 & 60 & 120 \\ 
Before & 337 & 48 & 97\\
Simultaneous & 288 & 41 & 83\\
Identity & 147 & 21 & 43 \\
Includes & 141 & 20 & 41 \\
IS\_included & 93 & 13 & 27 \\
Ended\_by & 66& 9 & 19 \\
During\_inv & 26 & 4 & 8 \\
Begun\_by  & 25 & 3 & 7 \\
Begins & 22 & 3 & 7 \\
IBefore & 16 & 2 & 5 \\
IAfter & 12 & 2 & 4 \\
During & 11 & 1 & 3 \\
Ends & 9 & 2 & 3 \\
\hline
Total & 1612 & 229 & 467 \\
\hline
\end{tabular}
\end{center}
\caption{\label{distribution} Distribution of temporal relations in TimeBank v1.2. }
\end{table}

We have used TimeBank corpus v1.2 for training and evaluating our model. The corpus consists of 14 temporal relations between 2308 event pairs, which are within the same sentence. These relations~\cite{sauri2006timeml} are {\it simultaneous, before, after, ibefore, iafter, begins, begun\_by, ends, ended\_by, includes, is\_included, during, during\_inv, identity}. Six pairs among them are inverse of each other and other two types are commutative ($e_1 \mathcal{R} e_2 \equiv e_2 \mathcal{R} e_1$, $\mathcal{R} \in $ {\it \{identical, simultaneous\}}). Our sequential model requires that relation should always be between $e_1$ and $e_2$, where $e_1$ occurs before $e_2$ in the sentence.  Therefore, before extracting the sequence, we inverted the relation types in cases where relation type was annotated in opposite order. Final distribution of dataset is given in Table \ref{distribution}.

\subsection{Extracting Context Word Sequence}

First, we extract words that are in the dependency path between two event mentions. However, event pairs can be very far in a sentence and are involved in complex syntactic structures. Therefore, we also apply two heuristic rules to deal with complex syntactic structures, e.g., two event mentions are in separate clauses and have a punctuation sign in their context. We describe our specific rules below. We used the Stanford parser~\cite{chen2014fast} for generating dependency relations and parts-of-speech tags and all notations follow enhanced universal dependencies~\cite{de2008stanford}. 

{\it Rule 1 (punctuation):} Comma directly influences the meaning in text and omitting it may alter the meaning of phrase. Therefore, include comma if it precedes or follows $e_1$, $e_2$ or their modifiers. 

{\it Rule 2 (children):} Modifiers like {\it now, then, will, yesterday, subsequent, when, was, etc.} contains information on the temporal order of events and help in grounding events to the timeline. These modifiers are often related to event mentions with a specific class of dependency relations. Include all such children of $e_1$, $ e_2$ and other words in the path between them, which are connected with dependency relations nmod:tmod, mark, case, aux, conj, expl, cc, cop, amod, advmod, punct, ref.

\subsection{Sequences and Classifier} 
We form three sequences on the extracted context words (with {\it t} words), which are based on (i) parts-of-speech tags: $\mathcal{P_T} = p_1, p_2,...,p_t$ (ii) dependency relations: $\mathcal{D_T} = d_1, d_2,..., d_n$\footnote{we only consider dependency relations for words in path connecting $e_1$ and $e_2$.} and (iii) word forms: $\mathcal{W_T} = w_1, w_2,..., w_t$. 

We transform each $p_i$ and $d_i$ to a one-hot vector and each $w_i$ to a pre-trained embedding vector ~\cite{pennington2014glove}. Then each sequence of vectors are encoded using their corresponding forward ($LSTM_f$) and backward ($LSTM_b$) LSTM layers.

\begin{figure}
  \centering \includegraphics [width=0.485\textwidth,keepaspectratio]{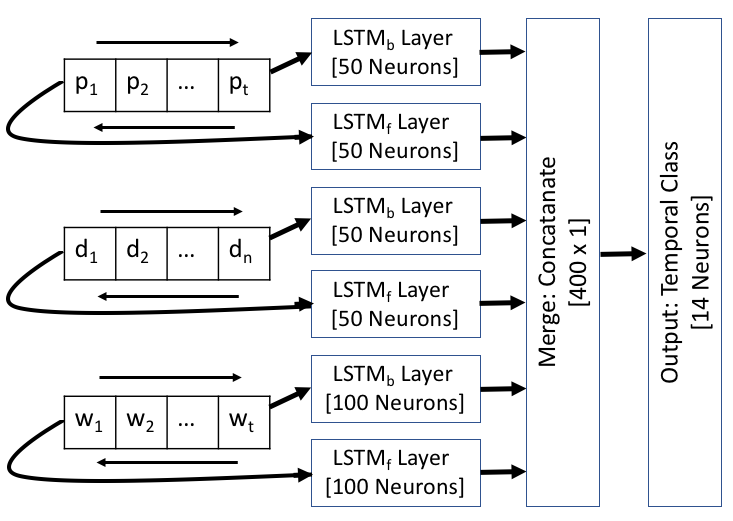}
  \caption{Bi-directional LSTM based classifier used for temporal relations classification.}
  \label{classifier}
\end{figure}

{\bf Classifier:} Figure \ref{classifier} shows an overview of our model. It consists of six LSTM~\cite{hochreiter1997long} layers, three of them encode feature sequences in forward order and remaining in reverse order. LSTM layers for POS tag and dependency relation have 50 neurons and have dropouts of 0.20. LSTM layers for word form have 100 neurons and have dropout of 0.25. All LSTM layers use 'tanh' activation function. Forward and backward embeddings of all sequences are concatenated and fed into another neural layer with 14 neurons corresponding to 14 fine-grained temporal relations. This neural layer uses softmax activation function. We train model for 100 iterations using rmsprop optimizer on batch size of 100 and error defined by categorical cross-entropy~\cite{chollet2015keras}
.

\section{Evaluation}
We evaluate our model using accuracy which has been used in previous research works for temporal relation classification. We also compare model performance using per-class F-score and macro F-score. We briefly describe all the systems we have used for evaluation.

{\it Majority Class}: assigns ``after'' relation to all event pairs.

{\it Unidirectional LSTMs}: use single LSTM layer to encode each sequence (POS tags, dependency relation and word forms) individually for extracted phrase in forward order.

{\it Bidirectional LSTMs}: use two LSTM layers to encode each sequence individually, taken from POS tags, dependency and word forms sequences. The first layer encodes sequence in forward and second in reverse order.

{\it 2 Sequences}: bi-directional LSTM based models considering all combinations of two sequences taken from POS tags, dependency and word forms sequences.

{\it Full model}: our complete sequential model considering POS, dependency and word forms sequences.

{\it Direct dependency path}: the same as {\bf Full model} except that the two heuristic rules were not applied in extracting sequences.

{\it Baseline I}: a neural network classifier using discrete features described in ~\citet{mirza2014classifying,d2013classifying}. The features used are: POS tag, dependency relation, token and lemma of $e_1$($e_2$); dependency relations between $e_1$($e_2$) and their children;  binary features indicating if $e_1$ and $e_2$ are related with the 'happens-before' or the 'similar' relation according to VerbOcean~\cite{chklovski2004verbocean},  if $e_1$ and $e_2$ have the same POS tag, or if $e_1$($e_2$) is the root and $e_1$ modifies (or governs) $e_2$; the dependency relation between $e_1$ and $e_2$ if they are directly connected in the dependency parse tree; prepositions that modify (or govern) $e_1$($e_2$); signal words ~\cite{derczynski2012using} and entity distance between $e_1$ and $e_2$. These features are concatenated and fed into an output neural layer with 14 neurons.

{\it Baseline II}: a neural network classifier using POS tags and word forms of words in the {\it surface path} as input. The {\it surface path} consists of words that lie in between two event mentions based on the original sentence. The classifier uses four LSTM layers to encode both POS tag and word sequences in forward and backward order. The output neural layer and parameters for all LSTM layers are kept the same as the {\it Full model}.

{\it Baseline III}: a neural network classifier based on event embeddings for both event mentions that were learned using bidirectional LSTMs \cite{kiperwasser2016simple}. The learning uses two LSTM layers, each with 150 neurons and dropout of 0.2, to embed the forward and backward representations for each event mention.
The input to LSTM layers are sequences of concatenated word embeddings and POS tags; each sequence corresponding to 19 context words to the left or to the right side of an event mention for the forward or the backward LSTM layer respectively. Event embeddings are then concatenated and fed into an output neural layer with 14 neurons.

All baselines are trained using rmsprop optimizer on an objective function defined by categorical cross entropy and their output layer uses softmax activation function.

\subsection{Results and Discussion}

\begin{table}[h]
\small
\begin{center}
\begin{tabular}{|l|l|l|}
\hline \bf Models & \bf $Accuracy$ \\ \hline
Majority Class & 25.69 \\ \hline
Baseline I & 41.97  \\ \hline
Unidirectional LSTM: only POS & 34.90  \\ 
\hspace{19.6ex} only Word & 35.12 \\
\hspace{19.6ex} only Dependency & 34.48 \\ \hline
Bidirectional LSTMs: only POS & 39.19  \\
\hspace{19.6ex} only Word& 37.69  \\ 
\hspace{19.6ex} only Dependency& 40.04  \\ \hline
2 Sequences: POS + Word& 44.54  \\
\hspace{10.2ex}Dependency + Word& 45.18  \\
\hspace{10.2ex}Dependency + POS& 47.75  \\ \hline
{\bf Full Model}& {\bf 53.32}  \\ \hline
Direct dependency path& 49.25  \\ 
\hline 
Baseline II& 43.90  \\ 
Baseline III& 44.75 \\ \hline
\end{tabular}
\end{center}
\caption{\label{results} Temporal relation classification result on TimeBank corpus.}
\end{table}

\begin{table}[h]
\small
\begin{center}
\begin{tabular}{|l|l|l|l|l|l|l|}
\hline
Relations & \multicolumn{3}{l|}{$Our System$} &  \multicolumn{3}{l|}{$Baseline I$} \\ 
&P&R&F&P&R&F\\ \hline
After & {\bf 0.62} & {\bf 0.68} & {\bf 0.65} & 0.56 & 0.48 & 0.45\\ 
Before & {\bf 0.56} & {\bf 0.52} & {\bf 0.53} & 0.37 & 0.45 & 0.41\\
Simultan. & {\bf 0.44} & {\bf 0.51} & {\bf 0.47} & 0.32 & 0.43 & 0.37\\
Identity & {\bf 0.47} & {\bf 0.56} & {\bf 0.51} & 0.45 & 0.53 &  0.49\\
Includes  & {\bf 0.59} & {\bf 0.39} & {\bf 0.47} & 0.43 & 0.30 & 0.35\\
IS\_includ. & 0.5 & {\bf 0.56} & 0.53 &{\bf 0.61} & 0.51 & {\bf 0.56}\\
Ended\_by & {\bf 0.48} & {\bf 0.63} & {\bf 0.55} & 0.41 & 0.47 & 0.44\\
During\_in. &  0 &  0 &  0 &0  & 0 &0\\
Begun\_by & {\bf 0.75}& {\bf 0.43} & {\bf 0.55}&0  & 0 &0 \\
Begins & {\bf 1.0} & {\bf 0.29} & {\bf 0.44} &0  & 0 &0\\
IBefore & {\bf 0.4} & {\bf 0.4} & {\bf 0.4} &0  & 0 &0\\
IAfter & {\bf 0.33} & {\bf 0.25} & {\bf 0.29} &0  & 0 &0\\
During & 0 & 0 & 0&0  & 0 &0 \\
Ends & 0 &  0 &  0 &0  & 0 &0\\
\hline
Macro Av. & {\bf 0.44} & {\bf 0.37} & {\bf 0.40} &0.23 & 0.22 & 0.22 \\
\hline
\end{tabular}
\end{center}
\caption{\label{fmacro} Per-class results of our best system and the baseline I.}
\end{table}

Table~\ref{results} reports accuracy scores for all the systems. We see that simple sequential models outperform the strong feature based system, {\it Baseline I}, which used various discrete features. Note that dependency relation and POS tag sequences alone achieve reasonably high accuracies. This implies that an important aspect of temporal relation is contained in the syntactic context of event mentions. Moreover, ~\citet{mirza2014classifying} observed that discrete features based on dependency parse tree did not contribute to improving their classifier's accuracy. On the contrary, using the sequence of dependency relations yields a high accuracy in our setting which signifies the advantages of using sequential representations for this task. Our {\it Full Model} achieves a performance gain of 11.35\% over {\it Baseline I}. 

We developed two more baselines ({\it Baseline II and III}) that do not require syntactic information as well as the {\it Direct dependency path} model that used no rules. The {\it Full Model} outperformed them by 9.42\%, 8.57\% and 4.07\% respectively. This affirms that the most useful syntactic and semantic structures are derived along dependency paths and additional context words, including prepositions, signal words and punctuations that are indirectly attached to event words, entail evidence on temporal relations as well.

Table~\ref{fmacro} compares precision, recall and $F_1$ scores of our {\it Full Model} with  {\it Baseline I}. Our model performs reasonably well compared to the baseline system for most of the classes. In addition, it is able to identify relations present in small proportion like {\it begun\_by, ibefore, iafter etc.}, which the baseline system couldn't identify. A similar observation was also reported by ~\citet{mirza2014classifying} that relation types {\it begins, ibefore, ends} and {\it during} are difficult to identify using feature based systems, which often generate false positives for {\it before} and {\it after} relations. 

\section{Conclusion and Future work}

In this paper, we have focused on modeling syntactic structural information and compositional semantics of contexts in predicting temporal relations between events in the same sentence. Our approach extracts lexical and syntactic sequences 
from contexts between two events and feed them to recurrent neural nets. The evaluation shows that our sequential models are promising in distinguishing among fine-grained temporal relations. 

In the future, we will extend our sequential models to predict temporal relations for event pairs spanning across multiple sentences, for instance by incorporating discourse relations between sentences in a sequence. 

\section*{Acknowledgments}

We want to thank our anonymous reviewers for their insightful review comments and suggestions that helped making evaluations more extensive.

\bibliography{emnlp2017}
\bibliographystyle{emnlp_natbib}

\end{document}